%% file: main.tex
\documentclass[conference]{IEEEtran}
\IEEEoverridecommandlockouts
\usepackage{booktabs}
\usepackage{cite}
\usepackage{amsmath,amssymb,amsfonts}
\usepackage{algorithmic}
\usepackage{graphicx}
\usepackage{textcomp}
\usepackage{xcolor}
\usepackage{pgfplots}
\usepackage{url}
\usepackage{hyperref}
\pgfplotsset{compat=1.17}
\usetikzlibrary{pgfplots.groupplots}
\usetikzlibrary{pgfplots.colormaps}
\usepgfplotslibrary{groupplots, colormaps, colorbrewer}
\usepackage{xcolor}
\usepackage{subcaption}

\pgfplotsset{
  cycle list/Paired-4   
}
\def\BibTeX{{\rm B\kern-.05em{\sc i\kern-.025em b}\kern-.08em
    T\kern-.1667em\lower.7ex\hbox{E}\kern-.125emX}}

\begin{document}

\title{\title{GeoToken: Hierarchical Geolocalization of Images via Next Token Prediction}}


\author{%
\IEEEauthorblockN{%
Narges Ghasemi\textsuperscript{*}\textsuperscript{\dag},
Amir Ziashahabi\textsuperscript{*}\textsuperscript{\ddag},
Salman Avestimehr\textsuperscript{\ddag},
Cyrus Shahabi\textsuperscript{\dag}}
\IEEEauthorblockA{\textsuperscript{\dag}\textit{Department of Computer Science}, University of Southern California, Los Angeles, CA, USA}
\IEEEauthorblockA{\textsuperscript{\ddag}\textit{Department of Electrical and Computer Engineering}, University of Southern California, Los Angeles, CA, USA}
\IEEEauthorblockA{\texttt{\{nghasemi, ziashaha, avestime, shahabi\}@usc.edu}}
\thanks{\textsuperscript{*} These authors contributed equally.}}


\maketitle

\makeatletter
\def\ps@IEEEtitlepagestyle{
  \def\@oddfoot{\mycopyrightnotice}
  \def\@evenfoot{}
}
\def\mycopyrightnotice{
  {\footnotesize
  \begin{minipage}{\textwidth}
  \centering
  ~\copyright~2025 IEEE. Personal use of this material is permitted. Permission from IEEE must be obtained for all other uses, in any current or future media, including reprinting/republishing this material for advertising or promotional purposes, creating new collective works, for resale or redistribution to servers or lists, or reuse of any copyrighted component of this work in other works.
  \end{minipage}
  }
}
\makeatother

\begin{abstract}

\input{text/abstract}
\end{abstract}


\input{text/introduction}

\input{text/related_work}
\input{text/method}

\input{text/results}

\input{text/conclusion}
\input{text/acknowledgments}

\bibliographystyle{IEEEtran} 
\bibliography{ref} 

\end{document}

%% file: text/abstract.tex
Image geolocalization—the task of determining an image's geographic origin—poses significant challenges, largely due to visual similarities across disparate locations and the large search space. To address these issues, we propose a hierarchical sequence prediction approach inspired by how humans narrow down locations from broad regions (e.g., country) to specific addresses (e.g., street name and house number). Analogously, our model predicts geographic tokens hierarchically, first identifying a general region and then sequentially refining predictions to increasingly precise locations. Rather than relying on explicit semantic partitions (e.g., country, city), our method uses S2 cells, a nested, multiresolution global grid, and sequentially predicts finer-level cells conditioned on visual inputs and previous predictions.
This procedure mirrors autoregressive text generation in large language models. Much like in language modeling, final performance depends not only on training but also on inference-time strategy. We investigate multiple top-down traversal methods for autoregressive sampling, incorporating techniques from test-time compute scaling used in language models. Specifically, we integrate beam search and multi-sample inference while exploring various selection strategies to determine the final output. This approach enables the model to manage uncertainty by exploring multiple plausible paths through the hierarchy.
We evaluate our method on the Im2GPS3k and YFCC4k datasets against two distinct sets of baselines: those that operate without a Multimodal Large Language Model (MLLM) and those that leverage one. In the MLLM-free setting, our model surpasses other comparable baselines on nearly all metrics, achieving state-of-the-art performance with accuracy gains of up to 13.9\%. When augmented with an MLLM, our model again outperforms all baselines, setting a new state of the art across every metric. The source code is available at \url{https://github.com/NNargesNN/GeoToken}.

%% file: text/introduction.tex

\section{Introduction}

Accurately estimating the geographic coordinates where a photograph was taken, a task known as worldwide image \textit{geolocalization}, is a long-standing problem in computer vision with broad practical applications. The ability to accurately geolocate images is crucial for organizing vast collections of visual data, enabling location-aware services in mobile applications, facilitating automatic photo organization, supporting environmental monitoring by connecting imagery to specific places for analysis, and enhancing search capabilities by allowing users to find photos taken in a particular area. Despite its utility, achieving accurate and robust worldwide image geolocalization remains a significant challenge.

There are two main challenges underlying this problem. Firstly, visual cues indicative of location are often subtle, ambiguous, and easily confounded. Features like architectural styles, vegetation types, or even road signs might offer hints, but similar elements can appear in different parts of the world, leading to potential confusion. Secondly, geotagged imagery, as a corpus for training or retrieval, is distributed unevenly across the globe. Figure \ref{fig:mp16-s2}(a) shows the distribution of the location of the images in one of the largest datasets used for training, MP-16 \cite{larson2017benchmarking} with over 4 million data points. 
 Popular tourist destinations and urban centers are often densely covered, while vast rural or remote areas have sparse or no associated geotagged images. This severe data imbalance means that models trained on existing datasets are heavily biased towards well-represented locations. This can lead to models that perform well in familiar areas but struggle to generalize and accurately predict locations in uncovered or underrepresented regions.

\newsavebox{\imgA}%
\newsavebox{\imgB}%
\sbox{\imgA}{\includegraphics[width=0.45\textwidth]{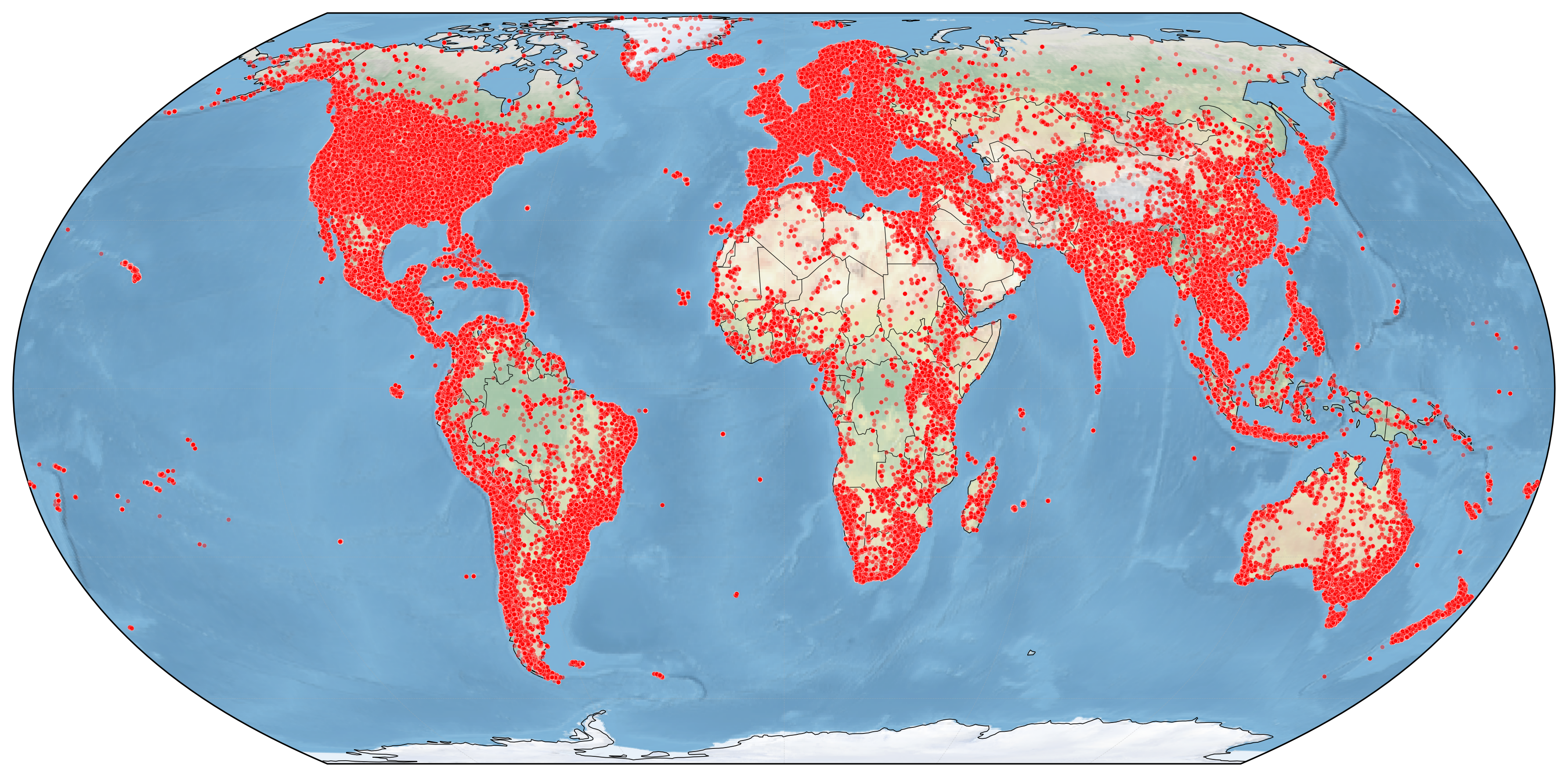}}
\sbox{\imgB}{\includegraphics[width=0.45\textwidth]{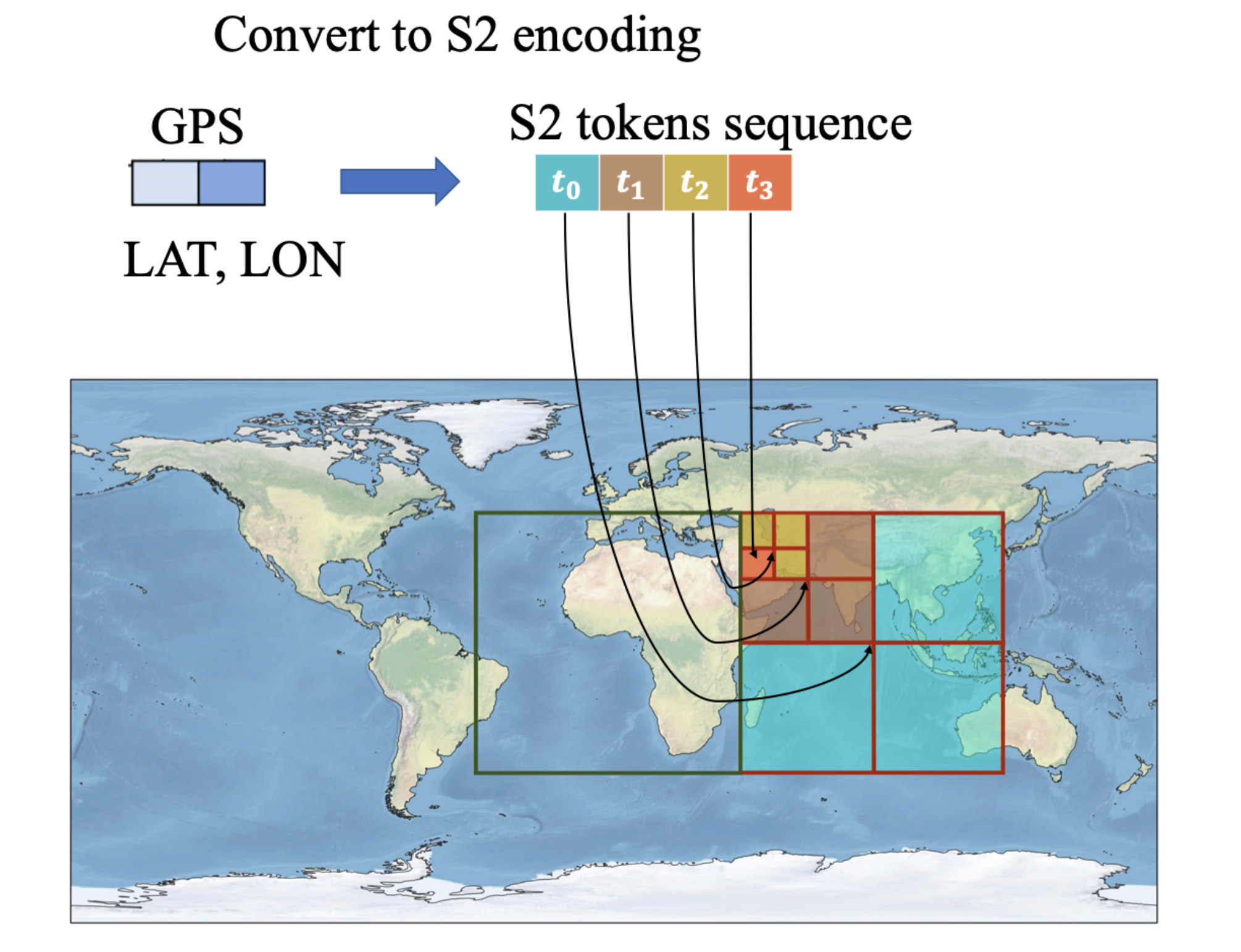}}
\begin{figure*}[htbp]
  \centering
  \usebox{\imgA}%
  \hfill
  \vrule width 1pt height \ht\imgA depth \dp\imgA%
  \hfill
  \usebox{\imgB}%
  \caption{%
    (a) Distribution of the MP16 dataset, with over 4 million samples across the world; 
    (b) Visualization of our S2 tokens.
  }
  \label{fig:mp16-s2}
\end{figure*}

Previous approaches to worldwide image geolocalization can be broadly categorized into three groups: classification, retrieval, and recent hybrid methods. Classification-based methods \cite{Weyand2016PlaNet, seo2018cplanet, Pramanick2022TransLocator, Clark2023Where} partition the Earth's surface into discrete geographic cells and train a model to predict which cell an image belongs to. While hierarchical structures were introduced to improve resolution \cite{Vo2017Revisiting}, a key limitation is their reliance on a relatively small number of predefined geographic cells; the model can only output one of the few discrete classes it was trained on, making it difficult to predict precise locations that are far from the center of the cells \cite{VivancoCepeda2023GeoCLIP}. Another fundamental strategy is image retrieval, where a query image is matched against a large database of geotagged images to find visually similar examples with known locations \cite{Vo2017Revisiting}. While effective for landmark-heavy datasets or areas with dense image coverage, retrieval struggles with the sheer scale of the Earth and the diversity of possible scenes; building a comprehensive global database is impractical, and many images, particularly in less photographed areas, will not have a close visual match \cite{Hays2008Im2GPS}. Furthermore, retrieval typically provides little sense of prediction confidence or uncertainty; it's difficult to gauge how reliable a retrieved location is, or if a visually similar match is truly geographically accurate or merely coincidental. More recently, hybrid methods and approaches leveraging Multimodal Large Language Models (MLLMs) or contrastive learning have emerged \cite{Zhou2023Img2Loc,Jia2024G3,Haas2024PIGEON,VivancoCepeda2023GeoCLIP}. These methods often combine aspects of retrieval and generation or learn powerful embeddings. While achieving impressive results, some heavily rely on closed-source models \cite{Zhou2023Img2Loc,Jia2024G3}, can involve complex pipelines \cite{Haas2024PIGEON}, and often lack an intuitive mechanism for managing prediction uncertainty.

In this paper, we introduce GeoToken, a novel approach that unifies the strengths of these diverse paradigms within a single, end-to-end framework, inspired by the coarse-to-fine reasoning process human experts employ when localizing an unfamiliar scene. Consider how a human might identify the location of a photo: they might first recognize broad regional cues ("This looks like the United States"), then refine their hypothesis based on more specific details ("The architecture suggests New York"), and finally pinpoint the city or street ("This building is clearly in Manhattan"). This process involves forming a broad initial hypothesis and progressively refining it as more evidence is considered.

Our core innovation is to translate this intuitive human strategy into a computational model by treating worldwide image geolocalization as a coarse-to-fine token prediction task. Analogous to how large language models \cite{radford2018improving, radford2019language,brown2020language,touvron2023llama} generate text, one token at a time. We capture this intuition by decomposing any geographic coordinate into a sequence of hierarchical tokens, where each token represents a progressively finer spatial subdivision. This process is analogous to reverse geocoding: translating a precise coordinate into a structured address composed of hierarchical components such as country, city, zipcode, street, and house number. Early tokens in our sequence correspond to broad regions, while later tokens refine the prediction to increasingly granular spatial detail. GeoToken then predicts this sequence of tokens autoregressively, predicting the next token for the next level in the hierarchy conditioned on the image and all previously predicted levels. This sequential generation process, traversing the geographic hierarchy, allows the model to build its location estimate incrementally.

To guide this process and provide robust spatial context, GeoToken integrates retrieval-augmented context. Taking inspiration from \cite{Jia2024G3}, we first train dual image-gps and image-text encoders, using a CLIP-style contrastive loss \cite{Radford2021Learning}. This ensures that we get image embeddings that naturally align with embeddings of their corresponding GPS coordinates and location descriptions, providing strong priors for both retrieval and generation.
Then, we use these encoders to retrieve context for our generation. Specifically, given an image input, we compute its embedding and use it to retrieve similar images from the training dataset. These retrieved images and their associated known token sequences serve as concrete "hints" to ground the generation process.

Furthermore, inspired by advancements in large language models, we leverage autoregressive decoding with test-time scaling techniques \cite{wang2022self, snell2024scaling}. This enables robust prediction and, crucially, inherently provides a natural way to manage uncertainty at each step of the prediction sequence. Utilizing techniques such as multi-answer sampling to explore alternative high-probability regions and maintain a set of plausible location hypotheses before committing to a final estimate allows us to extract robust predictions post-training. This mechanism of managing uncertainty mirrors human cognitive processes of considering alternatives and refining focus only when justified by strong evidence. A key benefit of this autoregressive, hierarchical approach and our training methodology is the ability to generate a rich pool of high-quality location candidates as the model explores the geographic hierarchy during decoding, offering a powerful alternative to the limited candidate sets from traditional methods. By framing the problem as a sequence generation over a geographic hierarchy, we move beyond fixed bins and retrieval limitations, offering a flexible and intuitive approach to worldwide geolocalization that mirrors human cognitive processes.

We validate the effectiveness of our approach through extensive experiments on the widely-used Im2GPS3k and YFCC4k benchmarks. To provide a comprehensive picture, we compare our performance in two distinct settings: one where our model operates independently (MLLM-free), and one where it is augmented by an MLLM. In the MLLM-free setting, our model surpasses other non-MLLM baselines on most metrics, achieving state-of-the-art performance with accuracy gains of up to 13.9\%. When augmented with an MLLM, our model again outperforms all baselines, achieving state-of-the-art performance across all metrics. Importantly, the strong performance in the MLLM-free setting unlocks the ability to perform highly accurate geolocalization entirely on-device, ensuring user data remains secure—a critical advantage over API-dependent methods.

In summary, our contributions are:  
\begin{itemize}
    \item We introduce a novel hierarchical sequence prediction framework for worldwide image geolocalization, drawing inspiration from human reasoning and autoregressive language modeling.
    \item We propose a context-guided autoregressive decoding mechanism, integrating retrieval-augmented context to enhance robustness across diverse locations.
    \item Our autoregressive generation process inherently supports sampling an unlimited number of guesses, allowing navigation of different hierarchical paths in cases of uncertainty. We show that performance can be improved by test-time scaling methods, through generating a high-quality pool of samples and leveraging selection strategies to derive the final answer.
    \item Empirical evaluations show state-of-the-art performance on nearly all metrics in MLLM-free setting, establishing a new overall state-of-the-art when augmented with MLLM.
\end{itemize}



%% file: text/related_work.tex
\section{Related Work}
Our work builds upon advances in three primary areas: the long-standing computer vision task of image geolocalization, the paradigm of autoregressive generation popularized by large language models, and the framework of retrieval-augmented generation for grounding predictions in external knowledge.

\subsection{Image Geolocalization}
The task of estimating a photo's geographic origin, known as image geolocalization, was pioneered by works like IM2GPS, which framed the problem as a large-scale image retrieval task \cite{Hays2008Im2GPS}. Since then, approaches have generally fallen into three main categories: classification, retrieval, and hybrid methods.

\paragraph{Classification-based Approaches.}

This paradigm partitions the Earth's surface into a discrete grid and classifies an image into one of the cells. PlaNet formulates localisation as a multiclass problem over $\sim$26000 cells \cite{Weyand2016PlaNet}. Hierarchical refinements—splitting coarse cells only where training data warrant—boost both resolution and sample efficiency (C-PlaNet \cite{seo2018cplanet}; ISNs \cite{muller2018geolocation}). Transformer backbones (TransLocator \cite{Pramanick2022TransLocator}) and vision-language pre-training (Clark et al. \cite{Clark2023Where}) further improve accuracy. While effective at a coarse level, these methods are limited by their predefined grid resolution and struggle to make precise, continuous predictions \cite{VivancoCepeda2023GeoCLIP}.

\paragraph{Retrieval-based Approaches.}
Concurrent to classification, retrieval-based methods match a query image against a large database of geotagged images, predicting the location of the best match or the average location of the top-$k$ matches \cite{Hays2008Im2GPS, arandjelovic2016netvlad}. These methods excel at recognizing specific landmarks but often fail in non-descript landscapes or regions with sparse data coverage in the retrieval gallery. Modern approaches have enhanced this paradigm with powerful deep learning features \cite{arandjelovic2016netvlad} and improved ranking strategies \cite{Vo2017Revisiting}.

\paragraph{Hybrid and Modern Approaches.}
Recently, the field has moved towards hybrid models that combine the strengths of both paradigms and leverage modern deep learning techniques. Contrastive learning has been used to learn joint embeddings for images and GPS coordinates, enabling both retrieval and direct regression-like prediction \cite{VivancoCepeda2023GeoCLIP}. Other works have introduced sophisticated losses on semantic cells \cite{Haas2024PIGEON} or have explicitly used large multimodal models (MLLMs) to interpret visual cues and retrieved context, treating geolocalization as a generative task \cite{Zhou2023Img2Loc, Jia2024G3}. Our work is inspired by this latter trend but focuses on a more fundamental sequence prediction approach that does not inherently depend on a pretrained MLLM for generation.

\subsection{Autoregressive Generation}
Autoregressive models are a fundamental class of generative models that produce complex, high-dimensional data one element at a time, where each new element is conditioned on all previously generated ones. This principle, which factorizes a joint distribution into a product of conditional probabilities, has long been the foundation of statistical language modeling \cite{bengio2003neural}.

The modern era of autoregressive generation was catalyzed by the introduction of the Transformer architecture \cite{vaswani2017attention}, which replaced recurrent networks with a more parallelizable self-attention mechanism. This architectural shift, combined with massive datasets and computational scale, led to the development of Large Language Models (LLMs) like the GPT series \cite{radford2019language, brown2020language}. These models demonstrated an unprecedented ability to generate coherent and contextually relevant text, performing a wide range of tasks in a zero-shot or few-shot manner.

More recently, this paradigm has been extended to multimodal contexts. Multimodal Large Language Models (MLLMs) like Flamingo \cite{alayrac2022flamingo} and LLaVA \cite{liu2023visual} condition the autoregressive text generation process on visual inputs, enabling them to describe images, answer visual questions, and perform complex reasoning over visual data. Our work adapts this core autoregressive principle, but instead of generating natural language, we generate a sequence of hierarchical geographic tokens conditioned on an image and its retrieved context.

\subsection{Retrieval-Augmented Generation (RAG)}
While large parametric models like LLMs store vast amounts of knowledge in their weights, they are prone to hallucination and their knowledge is static post-training. Retrieval-Augmented Generation (RAG) is a powerful framework designed to mitigate these issues by combining the parametric knowledge of a generator with a non-parametric external memory or database \cite{lewis2020retrieval}.

The standard RAG pipeline involves two stages. First, given a query, a retriever module searches a large corpus (e.g., Wikipedia) for relevant documents. Second, these retrieved documents are provided as additional context to a generator model (e.g., an LLM), which then produces the final output, grounded in the retrieved information. This approach has been shown to improve factuality, reduce hallucination, and allow for knowledge to be updated simply by changing the retrieval corpus, without costly retraining \cite{lewis2020retrieval}.

Initially developed for knowledge-intensive NLP tasks, the RAG concept has been extended to multimodal domains \cite{yasunaga2022retrieval}. In computer vision, this often involves retrieving relevant images, text, or other data to provide context for a given visual task. Recent geolocalization models such as G3 \cite{Jia2024G3} are prime examples of this trend, using retrieved images and their locations to inform a generative process. Our work builds directly on this paradigm, using a gallery of training images as our non-parametric memory and explicitly conditioning our autoregressive decoder on the retrieved context.

%% file: text/method.tex
\section{GeoToken}

\label{sec:method}

\begin{figure*}[!htb]
  \centering
  \includegraphics[width=\textwidth]{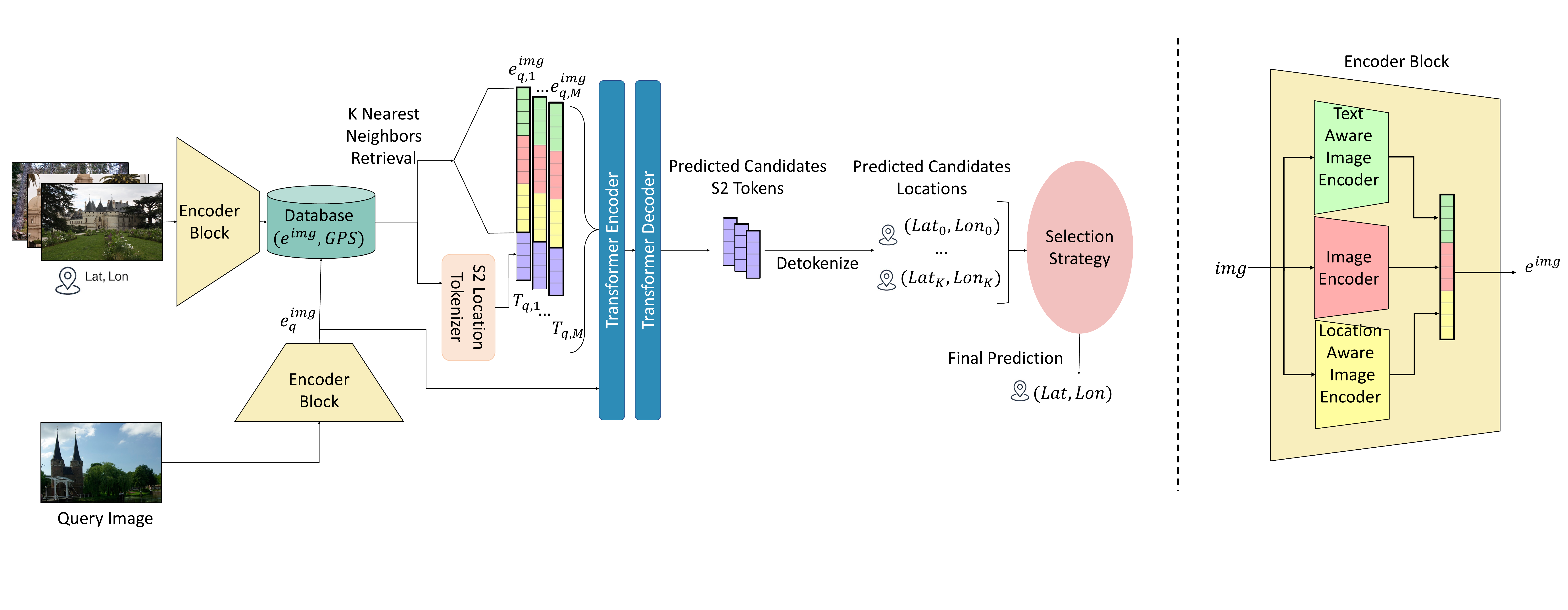}
  \caption{The GeoToken pipeline for retrieval-augmented geolocalization. A query image is encoded (1) and used to retrieve visually similar neighbors and their S2 location tokens from a gallery (2). This retrieved context grounds an encoder-decoder Transformer (3) that autoregressively predicts the final location as a hierarchical S2 token sequence. At test time, a pool of candidate locations is generated and a final prediction is chosen using a reranking strategy (4).}
  \label{fig:architecture}
\end{figure*}

An overview of GeoToken's architecture and inference pipeline is illustrated in Figure~\ref{fig:architecture}. Given a query image, GeoToken first generates a location-aware embedding using a pretrained encoder. This embedding serves two purposes: it acts as a primary input to the prediction model and is used to retrieve visually similar images and their known locations from a training gallery. These retrieved locations are then tokenized into a hierarchical sequence. Finally, the embeddings of the query image and its retrieved neighbors, along with the tokenized neighbor locations, are fed into a transformer model \cite{vaswani2017attention} that autoregressively predicts the query's location sequence. Optionally, various decoding and selection strategies can be employed to generate a pool of candidate locations and derive a final, robust prediction. The remainder of this section details each component of this framework.

\subsection{Location Representation with Hierarchical S2 Tokens}

As GeoToken predicts locations token-by-token in an autoregressive manner, we first require a method to convert geographic coordinates into a discrete sequence of tokens. To this end, we adopt Google’s S2 geometry for spatial indexing, which partitions the globe into a hierarchical quadtree structure.\footnote{\url{https://s2geometry.io/}} At the coarsest level (level 0), the Earth is projected onto a cube of six faces. Each cell is then recursively subdivided into four children (a quad-subdivision) to increase the resolution.

We represent a location as a sequence of $L$ tokens derived from its S2 representation, covering levels 0 through $L-1$. For our task, we use $L=21$, which provides precision down to a few hundred meters. This process converts a location's latitude and longitude into a token sequence:
\[
 T = [t_0, t_1, t_2, \dots, t_{L-1}],
\]
where $t_0 \in \{0, \dots, 5\}$ is the S2 face token (level 0), and each subsequent token $t_i \in \{0, 1, 2, 3\}$ for $i \in \{1, \dots, L-1\}$ encodes the quadrant at level $i$. This representation is inherently hierarchical; a shared prefix of length $l$ signifies that two locations fall within the same S2 cell at level $l-1$, providing an implicit notion of geographic proximity. For instance, locations within the same city will share a long common prefix, while those in different countries will diverge much earlier. This structure links token-space distance to real-world distance, as a small edit to a token sequence corresponds to moving to an adjacent region. The coarse-to-fine granularity also mirrors human-like descriptions, such as specifying a country, then a city, and finally a neighborhood.


\subsection{Encoder Pretraining via Geo-Alignment}
\label{subsec:geo_alignment}

GeoToken relies on powerful embeddings to provide informative inputs to the predictive model and to retrieve relevant context for generation. To obtain these embeddings, we follow the approach proposed by prior work G3 \cite{Jia2024G3} and train various encoders for encoding relevant information. This method learns expressive, location-aware representations by jointly aligning images with multi-modal geographical data: GPS coordinates and textual descriptions. Following G3, we define separate encoders for the image, GPS, and text modalities.

\subsubsection{Image Encoder}
For an input image $I_i$, a pretrained vision encoder $\mathcal{V}$ (e.g., ViT-L/14 \cite{dosovitskiy2020image} from CLIP \cite{Radford2021Learning}) first extracts raw visual features $e_i^{img\_raw} = \mathcal{V}(I_i)$. These features are subsequently projected into two distinct embedding spaces using trainable feed-forward networks, $f_{text}^{img\_proj}$ and $f_{gps}^{img\_proj}$, to facilitate alignment with textual and GPS modalities, respectively:
\begin{itemize}
    \item $e_{i}^{image\_text} = f_{text}^{img\_proj}(e_i^{img\_raw})$: Image embedding for alignment with textual location descriptions.
    \item $e_{i}^{image\_gps} = f_{gps}^{img\_proj}(e_i^{img\_raw})$: Image embedding for alignment with GPS coordinates.
\end{itemize}
For constructing a comprehensive image representation, embeddings are then concatenated to obtain final image embedding: $e_i^{image} = \text{concat}(e_i^{img\_raw}, e_{i}^{image\_text}, e_{i}^{image\_gps})$.

\subsubsection{GPS Coordinate and Text Encoders}
Following G3, we encode raw GPS coordinates using an encoder, which applies a Mercator \cite{snyder1987map} projection followed by multi-scale Random Fourier Features (RFF) \cite{tancik2020fourier} and feed-forward networks. Similarly, textual descriptions (e.g., ``Vestland, Norway''), obtained via reverse geocoding, are encoded using a pretrained text encoder and a trainable projection head.

\subsubsection{Geo-Alignment Training Objective}
These encoders are trained jointly with a symmetric contrastive loss (InfoNCE) \cite{oord2018representation}, which aligns the image embeddings with their corresponding GPS and text embeddings in a shared space. The total loss is:
\[
\mathcal{L}_{\text{GeoAlign}} = \frac{1}{2} (\mathcal{L}_{\text{image,text}} + \mathcal{L}_{\text{image,gps}} + \mathcal{L}_{\text{text,image}} + \mathcal{L}_{\text{gps,image}})
\]
This pretraining stage yields encoders that are highly attuned to location-indicative visual cues. For complete architectural details, we refer the reader to G3 \cite{Jia2024G3}.

\subsection{Retrieval-Augmented Generation}

With the encoders pretrained as described in Section~\ref{subsec:geo_alignment}, the model is ready for its primary task: retrieval-augmented generation. This process uses an encoder-decoder transformer architecture. The encoder first processes the query image alongside its retrieved context to produce a contextualized memory representation. The decoder then attends to this memory to autoregressively generate the final location sequence, token by token.

\paragraph{Context Retrieval}
The retrieval process begins with creating a gallery containing the image embedding $e_i^{\text{image}}$ for every image $i$ in the training dataset. As detailed in Section~\ref{subsec:geo_alignment}, this embedding concatenates raw visual features with specialized projections aligned to GPS and text data, providing a powerful, multi-aspect representation for search.

Given a query image $I_q$, we compute its embedding $e_q^{\text{image}}$ and use it to retrieve the top-$M$ nearest neighbors from the gallery via cosine similarity. For each neighbor, we retrieve both its image embedding and its ground-truth S2 token sequence.

\paragraph{Transformer Encoder Input}
The input to the transformer encoder is a sequence combining the query image with its retrieved context. Let $e_q^{\text{image}}$ be the query image embedding. For each of the $M$ nearest neighbors, let $e_{q,j}^{\text{image}}$ be its image embedding and $T_{q,j} = [T_{q,j}^{0}, \dots, T_{q,j}^{L-1}]$ be its S2 token sequence.

Each component is first projected into the transformer's hidden dimension $d$ using dedicated embedding layers: an image embedding layer $E_{\text{img}}(\cdot)$ and an S2 token embedding layer $E_{\text{tok}}(\cdot)$. Let the projected vectors be $v_q = E_{\text{img}}(e_q^{\text{image}})$, $v_{q,j} = E_{\text{img}}(e_{q,j}^{\text{image}})$, and $\mathbf{t}_{q,j} = E_{\text{tok}}(T_{q,j})$. The full input sequence for the encoder, $\mathbf{X}$, is then formed by concatenation:
\[
  \mathbf{X} = \Bigl[\, v_q \;\oplus\; \bigoplus_{j=1}^M \bigl( v_{q,j} \oplus \mathbf{t}_{q,j} \bigr) \,\Bigr],
\]
where $\oplus$ denotes concatenation along the sequence dimension. This yields a single sequence of length
\[
  1 + M \times (1+L),
\]
where each element is a vector in $\mathbb{R}^d$. Learned positional embeddings are added to differentiate the query from each neighbor's image and token block.

\paragraph{Autoregressive Decoder}
The encoder processes the input sequence $\mathbf{X}$ via self-attention to produce a contextualized memory representation, $\mathrm{Enc}(\mathbf{X})$. The causal transformer decoder then generates the S2 token sequence for the query, $(t_0,\dots,t_{L-1})$, one token at a time. At each step $s$, it models the conditional probability of the next token:
\[
  P\bigl(t_s \mid t_{<s},\,\mathrm{Enc}(\mathbf{X})\bigr).
\]
To do this, the decoder attends to the full output of the encoder, leveraging both the query's visual context and the retrieved geographic exemplars to inform its prediction for the next spatial token.

The decoder is effectively learning a language model over location tokens, conditioned on the rich context provided by the encoder. The autoregressive formulation models the joint probability of the sequence using the chain rule of probability:
\[
P(t_1, \dots, t_L \mid \mathrm{Enc}(\mathbf{X})) = \prod_{s=0}^{L-1} P(t_s \mid t_{<s}, \mathrm{Enc}(\mathbf{X})).
\]
This decomposition is well-suited for the task, as it allows the model to first focus on broad distinctions (the first few tokens) before gradually honing in on a precise location. If the input is ambiguous (e.g., between two neighboring cells at level 3), the model’s probability distribution at that step will reflect this uncertainty. The entire model is trained end-to-end with teacher forcing, using the a weighted sum of cross-entropy losses over all token positions as the objective.

\subsection{Decoding Strategies}
Generating a location from the trained model requires an autoregressive decoding process. The most straightforward approach, greedy decoding, selects the highest-probability token at each step. While fast, this deterministic method has two key weaknesses: it cannot represent uncertainty, forcing a single choice even when multiple regions are plausible, and it is susceptible to cascading errors, where an early mistake can derail the entire subsequent sequence.

To better handle ambiguity and improve robustness, we explore two families of decoding methods widely used in natural language processing: sampling with temperature and beam search.

\subsubsection{Sampling with Temperature}
Instead of deterministically picking the best token, we can sample from the probability distribution produced by the model at each step. This distribution is controlled by a temperature parameter $T$, which rescales the model's output logits $\boldsymbol{\ell}_t$:
\[
  \mathbf{p}_t = \mathrm{softmax}\!\Bigl(\frac{\boldsymbol{\ell}_t}{T}\Bigr).
\]
A temperature $T<1$ sharpens the distribution, favoring high-probability tokens, while $T>1$ flattens it, encouraging exploration. As $T \to 0$, sampling approaches greedy decoding. This controlled randomness is effective for exploring the solution space when the model is uncertain. By drawing multiple \textbf{independent} samples, we generate a set of plausible location sequences, increasing the probability that at least one candidate is highly accurate. Empirically, we find that a moderate temperature ($T \approx 0.5 \text{--} 0.7$) offers the best trade-off between reliability and diversity.

\subsubsection{Beam Search}
As a deterministic alternative, beam search maintains a "beam" of the top-$B$ most probable partial hypotheses at each step. It systematically expands each partial sequence and retains the $B$ new sequences with the highest cumulative log-probability:
\[
  \text{score}(\tau\oplus k) = \text{score}(\tau) + \log p(k\mid \tau),
\]
where $\tau$ is a partial sequence and $k$ is a candidate token. While beam search excels at finding high-probability sequences, its main drawback is its lack of stochasticity; an early, high-confidence error can trap the entire beam in an incorrect part of the search space.

\subsection{Candidate Reranking and Selection Strategies} \label{sec:candidate-reranking}
After generating a diverse pool of candidate sequences via decoding, a final step is required to derive the final prediction. We explore several strategies for this task.

\subsubsection{Log-Probability Selection}
The most direct method is to rely on the generative model's own confidence scores. Each candidate sequence $s$ has an associated cumulative log-probability:
\[
  \mathrm{score}(s) = \sum_{j=0}^{L-1} \log p\bigl(k_j \mid k_1, \dots, k_{j-1}\bigr).
\]
We select the candidate with the highest log-probability, which favors the sequence that the model itself considers most likely.

\subsubsection{Reward Model Reranking}
Another approach is to train a separate reward model to predict the accuracy of a given candidate sequence. To do this, we take a rather simple approach and discretize the continuous prediction error into binary bins corresponding to the evaluation distance thresholds $<$200\,km and $>$200\,km. The process is as follows:
\begin{enumerate}
  \item Generate Dataset: For each image in the training set, sample a set of candidate location sequences using our trained model.
  \item Label Data: For each candidate, decode its coordinates and compute its haversine distance to the ground truth, then label it with the corresponding error bin.
  \item Train Scorer: Train a classifier $c_{\psi}(s_i)$ on this dataset to predict the correct error bin for any given sequence $s_i$.
  \item Select Best: At inference time, apply the trained scorer to all candidates in the pool and select the one with the highest predicted probability of being in the smallest-error bin ($b=0$):
  \[
    s^* = \arg\max_i \bigl[c_{\psi}(s_i)\bigr]_{b=0}.
  \]
\end{enumerate}
\subsubsection{Similarity-based Selection}
This strategy leverages the shared embedding space learned during our geo-alignment pretraining (Section~\ref{subsec:geo_alignment}). For each candidate location sequence, we first decode it to get its GPS coordinate and then generate its corresponding location embedding using our pretrained GPS encoder. We then select the candidate whose location embedding exhibits the highest cosine similarity with the query image's embedding.

\subsubsection{MLLM-as-a-Judge} 
\label{ssec:mllm-as-judge}
This strategy employs a large multimodal model (MLLM) to arbitrate among the generated candidates. The MLLM is provided with the query image and the pool of candidates and can be used in one of two modes:
\begin{itemize}
  \item \textbf{Pool-Selection Mode:} The MLLM is prompted to choose the best option from a list of candidate coordinates.
  \item \textbf{Free-Generation Mode:} The MLLM is allowed to either pick one of the provided candidates or generate an entirely new coordinate if it determines none are sufficiently accurate.
\end{itemize}
The final prediction is then parsed from the MLLM's textual response.

%% file: text/results.tex
\begin{table*}[h]
  \centering
  \caption{Overall localization accuracy (\%) on IM2GPS3K and YFCC4K, with median error (km).}
  \label{tab:mllm-free}
\begin{tabular}{lccccc|ccccc}
\hline
Method  & \multicolumn{5}{c|}{IM2GPS3K} & \multicolumn{5}{c}{YFCC4K} \\
 &  1 km & 25 km & 200 km & 750 km & 2500 km & 1 km & 25 km & 200 km & 750 km & 2500 km \\
\hline
[L]kNN               &  7.2 & 19.4 & 26.9 & 38.9 & 55.9  &  2.3 &  5.7 & 11.0 & 23.5 & 42.0 \\
PlaNet                   &  8.5 & 24.8 & 34.3 & 48.4 & 64.6  &  5.6 & 14.3 & 22.2 & 36.4 & 55.8 \\
C-PlaNet                  & 10.2 & 26.5 & 34.6 & 48.6 & 64.6  &  7.9 & 14.8 & 21.9 & 36.4 & 55.5 \\
ISN                       & 10.5 & 28.0 & 36.6 & 49.7 & 66.0  &  6.5 & 16.2 & 23.8 & 37.4 & 55.0 \\
TransLocator              & 11.8 & 31.1 & 46.7 & 58.9 & 80.1  &  8.4 & 18.6 & 27.0 & 41.1 & 60.4 \\
Clark et al.              & 12.8 & 33.5 & 45.9 & 61.0 & 76.1  & 10.3 & \underline{24.4} & 33.9 & 50.0 & 68.7 \\
GeoCLIP                   & \underline{14.1} & 34.5 & 50.7 & 69.7 & 83.8  &  9.6 & 19.3 & 32.6 & 55.0 & 74.7 \\
PIGEON                    & 11.3 & \underline{36.7} & \bf{53.8} & \bf{72.4} & \bf{85.3}  & \underline{10.4} & 23.7 & \underline{40.6} & \underline{62.2} & \underline{77.7} \\
\hline

GeoToken &      \bf{16.8}  &  \bf{39.6}  &  \bf{53.8}  &  \underline{70.8}  &  \underline{85.0}  &  \bf{24.3}  &  \bf{35.3}  &  \bf{46.6}  &  \bf{64.2}  &  \bf{78.6}    \\
\hline
\end{tabular}
\end{table*}


\begin{table*}[h]
  \centering
  \caption{Comparison of Localization accuracy (\%) using GeoToken, Img2Loc, and G3 under the MLLM‐assisted setting using Gemini‐2.0‐Flash on IM2GPS3K and YFCC4K. }
  \label{tab:mllm-assisted}
\begin{tabular}{lccccc|ccccc}
\hline
Method & \multicolumn{5}{c|}{IM2GPS3K} & \multicolumn{5}{c}{YFCC4K} \\
 & 1 km & 25 km & 200 km & 750 km & 2500 km & 1 km & 25 km & 200 km & 750 km & 2500 km \\
\hline
Img2Loc               & 16.4 & 42.5 & 55.6& 72.2 & 85.3  & 18.7 & 31.6 & 43.8 & 62.0 & 76.1 \\

G3                    & 17.2 & 44.4 & 59.1 & 74.6 & 86.8  & 22.9 & 37.2 & \underline{50.3} & 66.9 & 79.9 \\
\hline

GeoToken  (Pool-Selection Mode)  &  \underline{18.8}  &  \underline{45.0}  &  \underline{59.3}  &  \underline{75.2}  &  \underline{87.7}  &
              \underline{24.7}  &  \underline{37.7}  &  \underline{50.3}  &  \underline{67.0}  &  \underline{80.5}   \\
              
GeoToken (Free-Generation Mode) &  \bf{19.0}  &  \bf{46.0}  &  \bf{60.1}  &  \bf{76.6}  &  \bf{88.8}  &
              \bf{25.4}  &  \bf{38.5}  &  \bf{51.4}  &  \bf{68.0}  &  \bf{81.0}   \\

\hline
\end{tabular}
\end{table*}


\section{Experiments}

In this section, we detail a comprehensive evaluation designed to validate our hierarchical sequence prediction approach. We benchmark GeoToken against a wide array of state-of-the-art methods on two standard geolocalization datasets: IM2GPS3K and YFCC4K. 

\subsection{Experimental Setup}
\label{sec:setup}

\subsubsection{Datasets}

Our experiments leverage one training corpus and two distinct evaluation benchmarks to assess both in-distribution and out-of-domain generalization.

\begin{itemize}
    \item \textbf{MP16-Pro (Training):} This is our large-scale training corpus, derived from the original MP16 dataset. It contains approximately 4.1 million Flickr images. Following the procedure in \cite{Jia2024G3}, each image is annotated with multi-level geographic text (e.g., city, country, continent) using Nominatim. This dataset is used for both the initial CLIP-style geo-alignment and for training the main GeoToken model.
    \item \textbf{IM2GPS3K (Evaluation):} This benchmark contains 3,000 diverse, globally-distributed images and is a strong test of out-of-domain generalization \cite{Hays2008Im2GPS}. Its emphasis on rural and non-landmark scenes makes it particularly challenging for retrieval-based methods.
    \item \textbf{YFCC4K (Evaluation):} This benchmark is a 4,000-image subset of the YFCC100M dataset \cite{thomee2016yfcc100m}. In contrast to IM2GPS3K, its distribution of urban scenes and popular landmarks more closely mirrors our training data, testing the model's performance on more familiar-looking scenes.
\end{itemize}

\subsection{Baselines}

We compare GeoToken with the most widely-used models in image geolocalisation:

\begin{itemize}
  \item \textbf{$k$-NN} (\emph{$\sigma{=}4$}) \cite{Vo2017Revisiting}.\;
        Returns the mean location of the $k$ nearest visual neighbours;  
        shrinking $k$ lowers the Gaussian bandwidth and approaches plain 1-NN. 
        
  \item \textbf{PlaNet} \cite{seo2018cplanet}.\;
        Casts the task as a single multi-class classification problem by partitioning the globe into thousands of cells.

  \item \textbf{C-PlaNet} \cite{seo2018cplanet}.\;
        Improves PlaNet by letting overlapping coarse cells vote for finer-grained intersections.

  \item \textbf{ISNs} \cite{muller2018geolocation}.\;
        Adds a parallel scene-context branch (indoor, urban, natural) and fuses it with hierarchical cell scores.

  \item \textbf{TransLocator} \cite{pramanick2022world}.\;
        Processes the raw image and its semantic-segmentation map through a dual-stream transformer.

  \item \textbf{GeoDecoder} \cite{clark2023we}.\;
        Applies cross-attention between coarse and fine tokens to reduce error propagation in deep hierarchies.

  \item \textbf{GeoCLIP} \cite{VivancoCepeda2023GeoCLIP}.\;
        Learns a GPS encoder that aligns CLIP image embeddings with location vectors.

  \item \textbf{Img2Loc} \cite{zhou2024img2loc}.\;
        Treats geolocalisation as retrieval-augmented generation: retrieved coordinates become tokens in an MLLM prompt.

  \item \textbf{PIGEON / PIGEOTTO} \cite{Haas2024PIGEON}.\;
        Creates semantic geo-cells and introduces a distance-aware smoothing loss that softens class boundaries.

  \item \textbf{G3} \cite{Jia2024G3}.\;
        Combines large-scale retrieval with a generative prior, drawing several candidate coordinates before a final selection step using the similarity-based approach.
\end{itemize}

\subsection{Implementation Details}
\label{ssec:impl}

\paragraph{Hierarchical S2 Tokenization} Every latitude-longitude coordinate is converted into a 21-token sequence using Google's S2 geometry library at level 20. This sequence consists of one token for the initial cube face (from a vocabulary of 6) and 20 subsequent quad-tree tokens (each from a vocabulary of 4) that progressively refine the location. A single embedding table is shared across all 21 positions.

\paragraph{Model Architecture} GeoToken is a 10-layer encoder-decoder Transformer ($d_{\mathrm{model}}{=}512$, 8 attention heads, 1024-dim FFN). It processes a concatenated sequence of specialized embeddings derived from a frozen CLIP ViT-L/14 backbone. The input consists of a learnable \texttt{[CLS]} token followed by projections representing the query image, its ground-truth location and text metadata, and context from its top-15 retrieved neighbors from the MP16-Pro gallery.

\paragraph{Training} Training proceeds in two stages. First, we perform Geo-Alignment by training the image and location encoders with a symmetric InfoNCE loss for 10 epochs to align their embeddings. Second, the full GeoToken model is trained for 50 epochs on MP16-Pro using AdamW (initial LR $5\times10^{-5}$, weight decay $10^{-6}$). Batches of 2048 are trained on a single NVIDIA GH200 GPU. The loss function is a position-weighted cross-entropy (CE) that penalizes errors at coarser levels of the S2 hierarchy more heavily:
\begin{equation*}
    \mathcal{L}= \frac{1}{\sum_{t}w_{t}} \sum_{t=0}^{20} w_{t}\, \mathrm{CE}(\hat y_{t},\,y_{t}),\quad \text{where } w_{t}=2.0-\frac{t}{20}
\end{equation*}



\paragraph{Evaluation Protocol.}
We report accuracy at standard distance thresholds (\{1, 25, 200, 750, 2500\}\,km) and median geodesic error. Our experiments use three general evaluation protocols to assess different aspects of our framework:
\begin{enumerate}
    \item Single Deterministic Prediction: Evaluates a single output from the model, produced by a deterministic decoding strategy like greedy decoding or beam search. This protocol is used to assess the core model's performance against MLLM-free baselines.

    \item Selected-from-Pool Prediction: Evaluates the final prediction after a selection strategy is applied to a pool of generated candidates (typically $K=30$ candidates from temperature sampling). The specific selection strategy varies by experiment and includes our main MLLM-as-a-Judge pipeline as well as other methods analyzed in our ablations (e.g., log-probability, similarity-based).

    \item Candidate Pool Quality (Ideal Selector): Measures the upper-bound potential of our generative model by reporting the accuracy of the best possible candidate within the generated pool (i.e., the closest-in-pool). This is used in our ablation studies to analyze the quality of the candidate set itself, independent of the selection strategy.
\end{enumerate}

\begin{table*}[h!]
  \centering
  \caption{Ablation of candidate selection strategies (\%) on IM2GPS3K and YFCC4K.}
  \label{tab:ablation_selection}
\begin{tabular}{lccccc|ccccc}
\hline
Method & \multicolumn{5}{c|}{IM2GPS3K} & \multicolumn{5}{c}{YFCC4K} \\
 & 1 km & 25 km & 200 km & 750 km & 2500 km & 1 km & 25 km & 200 km & 750 km & 2500 km \\
\hline

Ideal Selector &     33.1 & {59.2}  &  {77.3}      &  {90.1}  &  {95.7}  &  {39.1}  &  {56.3}  &  {75.2}  &  {89.9}  &  {96.3}    \\
\hline

    Log‐Probability  
      & 16.4 & 38.6 & 52.1 & 68.6 & 83.2 
      & 26.1 & 36.6 & 47.2 & 63.9 & 78.5  \\
        Beam Search (beam=2)  
      &  16.9 & 39.7 & 53.3 & 69.9 & 84.2 
      &  25.8 & 36.5 & 47.4 & 64.5 & 78.4\\

           Beam Search (beam=3)    
      &  16.2 & 38.7 & 52.7 & 69.2 & 83.7 
      & 26.2 & 36.4 & 47.4 & 64.2 & 78.7 \\

           Beam Search (beam=4)    
      &  15.7 & 38.0 & 51.9 & 69.0 & 83.2
      & \textbf{26.5} & 36.7 & 47.5 & 64.3 & 78.6 \\
      
    Reward Model (bin‐classifier)  
      &  14.5 & 35.1 & 48.0 & 65.5& 79.9 & 19.0 & 29.5 & 42.2& 60.0 & 74.9
       \\

    CLIP Similarity  
      & 14.1 & 36.7 & 51.7 & 69.9 & 83.7  
      &19.0 & 30.2 & 43.5 & 61.4 & 77.0  \\
    MLLM-as-a-Judge (Pool‐Selection)  
       &  18.8  &  45.0  &  59.3  &  75.2  &  87.7  &
              24.7  &  37.7  &  50.3  &  67.0  &  80.5   \\
    MLLM-as-a-Judge (Free‐Generation)  &
      \bf{19.0}  &  \bf{46.0}  &  \bf{60.1}  &  \bf{76.6}  &  \bf{88.8}  &
              25.4  &  \bf{38.5}  &  \bf{51.4}  &  \bf{68.0}  &  \bf{81.0}   \\

    \bottomrule
  \end{tabular}
\end{table*}

\definecolor{myblue}{RGB}{31,119,180}
\definecolor{myorange}{RGB}{255,127,14}
\definecolor{mygreen}{RGB}{44,160,44}
\definecolor{myred}{RGB}{214,39,40}

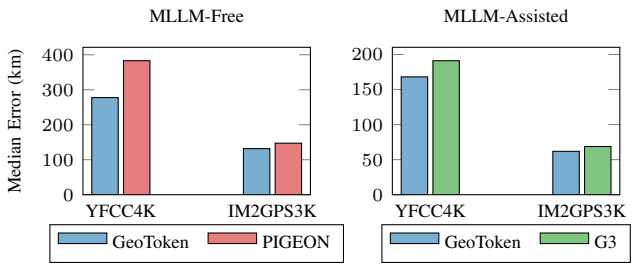
\begin{figure}[!t]
  \centering
  \begin{tikzpicture}
    \begin{groupplot}[
      group style={
        group size=2 by 1,
        horizontal sep=3.1em        
      },
      width=0.52\columnwidth,
      height=0.40\columnwidth,
      ybar,
      enlarge x limits=0.25,
      ymin=0,
      ylabel={Median Error (km)},
      ylabel style={font=\scriptsize},
      symbolic x coords={YFCC4K, IM2GPS3K},
      xtick=data,
      xtick style={draw=none},
      tick align=inside,
      tick label style={font=\scriptsize},
      xlabel style={font=\scriptsize},
      title style={font=\scriptsize},
      legend style={
        font=\scriptsize,
        at={(0.5,-0.20)},
        anchor=north,
        /tikz/every even column/.append style={column sep=0.5em}
      },
      legend columns=2,
      legend image code/.code={
        \draw[#1, draw=black] (0,0) rectangle (0.4em,0.4em);
      }
    ]

      \nextgroupplot[
        title={MLLM-Free},
        xlabel={Dataset}
      ]
        \addplot[fill=myblue!60!white] coordinates {(YFCC4K,277.559) (IM2GPS3K,131.621)};
        \addplot[fill=myred!60!white]  coordinates {(YFCC4K,383.000) (IM2GPS3K,147.300)};
        \legend{GeoToken, PIGEON}

      \nextgroupplot[
        title={MLLM-Assisted},
        ylabel={},      
        xlabel={Dataset}
      ]
        \addplot[fill=myblue!60!white]  coordinates {(YFCC4K,168.050) (IM2GPS3K,61.890)};
        \addplot[fill=mygreen!60!white] coordinates {(YFCC4K,191.110) (IM2GPS3K,68.740)};
        \legend{GeoToken, G3}

    \end{groupplot}
  \end{tikzpicture}
  \caption{Comparison of the median localization error (km) on YFCC4K and IM2GPS3K of GeoToken and prior state-of-the-art approaches.
           \textbf{Left}: MLLM-Free (GeoToken vs.\ PIGEON). 
           \textbf{Right}: MLLM-Assisted (GeoToken vs.\ G3).}
  \label{fig:median-error-comparison}
\end{figure}

\subsection{Main Quantitative Results}
\label{ssec:overall-results}
Baseline methods fall into two categories: (1) those that work locally and do not rely on a powerful MLLM for prediction (MLLM-free), and (2) those that leverage MLLMs (MLLM-assisted). To ensure fair comparisons and to demonstrate the performance of GeoToken as a standalone model, we separate our evaluations into these two settings. In the MLLM-free setting, we evaluate a single greedy decoding prediction from GeoToken against comparable baselines. In the MLLM-assisted setting, we compare our full pipeline (sampling 30 candidates with an MLLM-as-a-Judge) against G3 and Img2Loc. For a direct comparison, we reproduce the results for these baselines using the same MLLM judge (Gemini 2 Flash) as our own method. Our reproduced results for these baselines are generally higher than those originally reported, confirming a fair and strong comparison.

As shown in Table~\ref{tab:mllm-free}, in the MLLM-free setting, GeoToken’s greedy prediction establishes a new state of the art on nearly all metrics across both benchmark datasets. On the challenging IM2GPS3K dataset, GeoToken significantly improves accuracy at finer scales (e.g., 1\,km and 25\,km), where prior methods often struggle. The improvement is even more pronounced on YFCC4K, where at 1\,km GeoToken more than doubles the accuracy of the next-best method and maintains its lead across all subsequent radii.

In the LLM-assisted setting (Table~\ref{tab:mllm-assisted}), GeoToken's performance is further amplified. It consistently outperforms other MLLM-augmented methods like Img2Loc and G3 across all distance thresholds on both datasets. These results underscore the strength of our hierarchical generation approach, which provides a superior candidate pool for the MLLM judge to refine. As shown in Figure~\ref{fig:median-error-comparison}, these accuracy gains are reflected in a lower median prediction error compared to prior art in both evaluation settings.

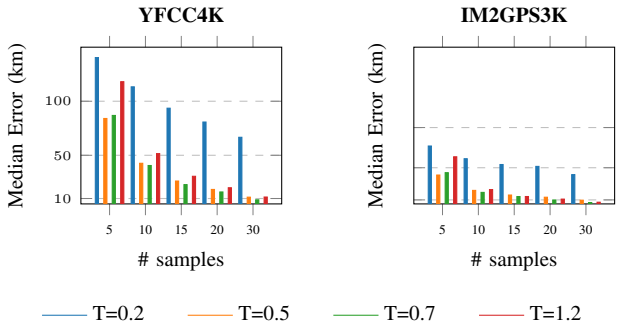
\begin{figure}[!t] 
\centering

\begin{subfigure}[t]{0.48\columnwidth}
    \centering
    \begin{tikzpicture}
        \begin{axis}[
            width=\linewidth, 
            title={\textbf{YFCC4K}},
            title style={font=\footnotesize},
            ybar,
            ymode=normal,
            ylabel={Median Error (km)},
            symbolic x coords={5, 10, 15, 20, 30},
            xtick=data,
            xlabel={\# samples },
            label style={font=\footnotesize},
            tick label style={font=\tiny},
            bar width=1.2 pt,
            enlarge x limits=0.2,
            ymin=5, ymax=150,
            ytick={10, 50, 100},
            ymajorgrids=true,
            grid style=dashed,
        ]
        \addplot[style={myblue, fill=myblue, mark=none}] coordinates {(5, 140.37) (10, 113.33) (15, 93.57) (20, 80.77) (30, 66.66)};
        \addplot[style={myorange, fill=myorange, mark=none}] coordinates {(5, 84.05) (10, 42.60) (15, 26.18) (20, 18.40) (30, 11.26)};
        \addplot[style={mygreen, fill=mygreen, mark=none}] coordinates {(5, 86.86) (10, 40.54) (15, 22.85) (20, 16.06) (30, 8.80)};
        \addplot[style={myred, fill=myred, mark=none}] coordinates {(5, 118.07) (10, 51.53) (15, 30.57) (20, 19.96) (30, 11.38)};
        \end{axis}
    \end{tikzpicture}
\end{subfigure}%
\hfill 
\begin{subfigure}[t]{0.48\columnwidth}
    \centering
    \begin{tikzpicture}
        \begin{axis}[
            width=\linewidth,
            title={\textbf{IM2GPS3K}},
            title style={font=\footnotesize},
            ybar,
            ylabel={Median Error (km)},
            ymode=normal,
            symbolic x coords={5, 10, 15, 20, 30},
            xtick=data,
            yticklabels={}, 
            xlabel={\# samples},
            label style={font=\footnotesize},
            tick label style={font=\tiny},
            bar width=1.2 pt,
            enlarge x limits=0.2,
            ymin=5, ymax=200,
            ytick={10, 50, 100},
            ymajorgrids=true,
            grid style=dashed,
        ]
        \addplot[style={myblue, fill=myblue, mark=none}] coordinates {(5, 77.05) (10, 61.23) (15, 54.03) (20, 51.70) (30, 41.44)};
        \addplot[style={myorange, fill=myorange, mark=none}] coordinates {(5, 40.96) (10, 21.84) (15, 16.07) (20, 13.34) (30, 9.66)};
        \addplot[style={mygreen, fill=mygreen, mark=none}] coordinates {(5, 43.86) (10, 19.35) (15, 14.18) (20, 9.91) (30, 6.72)};
        \addplot[style={myred, fill=myred, mark=none}] coordinates {(5, 63.63) (10, 22.81) (15, 14.22) (20, 10.99) (30, 7.12)};
        \end{axis}
    \end{tikzpicture}
\end{subfigure}

\par\vspace{0.5\baselineskip}
\begin{tikzpicture}
    \begin{axis}[
        hide axis,
        xmin=0, xmax=4, ymin=0, ymax=0,
        legend style={
            at={(0.5,0)}, 
            anchor=north,
            legend columns=-1,
            draw=none, 
            /tikz/every even column/.append style={column sep=0.5cm},
            font=\footnotesize,
            row sep=0pt, 
        },
    ]
    \addlegendimage{myblue,fill}
    \addlegendimage{myorange,fill}
    \addlegendimage{mygreen,fill}
    \addlegendimage{myred,fill}
    \legend{T=0.2, T=0.5, T=0.7, T=1.2}
    \end{axis}
\end{tikzpicture}
\caption{Median error of best closest-in-pool using different numbers of samples and temperatures on both datasets.}
\label{fig:best-guess-median}
\end{figure}

\subsection{Ablation Studies}
\subsubsection{Effect of Sample Pool Size and Temperature}
To understand how the quality of the candidate pool is affected by the number of candidates ($k$) and the sampling temperature ($T$), we perform a “closest-in-pool” ablation on both YFCC4K and IM2GPS3K. We generate 30 candidate sequences per image and vary:
\begin{itemize}
    \item $k \in \{5, 10, 15, 20, 30\}$, the number of sampled candidates per image.
    \item $T \in \{0.2, 0.5, 0.7, 1.2\}$, the sampling temperature.
\end{itemize}
Figure~\ref{fig:best-guess-median} illustrates the median error of the best guess in the candidate pool as a function of $k$ for each temperature. Key observations include: (1) As expected, the median error considerably decreases as $k$ increases, showing the benefit of a larger pool. (2) Intermediate temperatures ($T=0.5$ or $T=0.7$) perform best, balancing exploration and exploitation. (3) YFCC4K exhibits larger median errors than IM2GPS3K across all settings due to higher scene diversity. Based on these results, we adopted $T=0.7$ and $k=30$ as default sampling hyperparameters.

\subsubsection{Effect of Candidate Selection Strategy}
In our main results, we use MLLM-as-a-Judge as our default selection strategy. Here, we compare it against beam search and other methods detailed in Section~\ref{sec:candidate-reranking}. For each query, we generate 30 candidates ($T=0.7$) and apply different selection strategies. Table~\ref{tab:ablation_selection} shows the results. Ranking by log-probability and beam search provide strong performance, outperforming greedy prediction on many metrics. In contrast, CLIP similarity and the Reward Model perform poorly. The MLLM-as-a-Judge approaches offer a significant boost, with free-generation being the most effective. However, a significant gap remains between our best selector and the "Ideal Selector" (which would always pick closest-in-pool). This indicates that the candidate pool contains significantly higher-quality samples than our current selection strategies can identify, suggesting great potential for future work on more powerful selection mechanisms.

\section{Discussion: Private inference}

Unlike methods that rely heavily on MLLM APIs for their core prediction, GeoToken's architecture is able to provide strong performance in its MLLM-free configuration. This has critical privacy implications. Because all inference can be performed locally without sending data to a third party, making GeoToken suitable for on-device or private-server applications. Users can geolocate their images, which may contain sensitive personal information, without exposing them to the risks associated with external cloud services. This ``local-first'' capability ensures users retain full control over their data, a significant advantage over other leading models.

%% file: text/conclusion.tex
\section{Conclusion}
\label{sec:conclusion}

Inspired by advances in large language modeling and hierarchical decoding, we have presented GeoToken, a hierarchical sequence‐prediction framework that mirrors human coarse‐to‐fine reasoning for worldwide image geolocalization. By treating location as a sequence of S2‐cell tokens, our model first narrows down broad regions and then refines its prediction step by step, allowing it to capture both high‐level and fine‐grained geographic cues.  This autoregressive setup not only achieves state‐of‐the‐art accuracy without any external LLM, outperforming all non‐MLLM baselines by large margins at nearly every distance threshold, but also naturally supports sampling multiple plausible location hypotheses at inference time.  

Sampling from GeoToken provides two key benefits.  First, generating a pool of candidate coordinates lets the model explicitly manage uncertainty: if the visual evidence is ambiguous, multiple hierarchical paths can be explored before committing. Second, this sampling process is entirely local, preserving user privacy by avoiding reliance on external retrieval services or cloud‐based APIs. When an MLLM judge is added to select or refine among these candidates, GeoToken further extends its lead, even over other LLM‐augmented pipelines, demonstrating that our hierarchical predictions remain superior whether or not an LLM is used.  

Looking forward, GeoToken’s flexibility invites easy integration of additional modalities (e.g., timestamps, low‐resolution satellite imagery) or more advanced retrieval schemes, further enhancing robustness in underrepresented regions.  By releasing our code, pretrained weights, and the MP16-Pro splits, we hope to encourage future work that leverages hierarchical token prediction and uncertainty‐aware sampling to push the frontiers of worldwide image geolocalization.  

%% file: text/acknowledgments.tex
\section{Acknowledgments}
This research has been funded in part by NSF grants IIS-2128661 and 1956435, and NIH grant
5R01LM014026. Opinions, findings, conclusions, or recommendations
expressed are those of the author(s) and do not necessarily reflect the views of any sponsors, such as NSF or NIH.